\documentclass[10pt,conference,letterpaper]{IEEEtran}
\IEEEoverridecommandlockouts
\usepackage{cite}
\usepackage{amsmath,amssymb,amsfonts}
\usepackage{graphicx}
\usepackage{url}
\usepackage{balance}
\usepackage[hidelinks]{hyperref}
\usepackage{xcolor}

\newcommand{\toprule}{\hline}
\newcommand{\midrule}{\hline}
\newcommand{\bottomrule}{\hline}

\hypersetup{pdftitle={Detector Confidence Signals Presence Rather Than Occlusion in Cluttered Manipulation},pdfauthor={Yuanzhi He}}

\begin{document}

\title{Detector Confidence Signals Presence Rather Than\\ Occlusion in Cluttered Manipulation\thanks{The paper is accepted for IEEE MSN 2026 conference.}}

\author{\IEEEauthorblockN{Yuanzhi He}
\IEEEauthorblockA{School of Computational and Mathematical Sciences\\
Cardiff University, United Kingdom}}

\maketitle

\begin{abstract}
Detector confidence is often used to decide whether a robot needs another view of a target. We test this signal against detector-free visibility from geometry segmentation. In controlled clutter, Grounding DINO confidence remains nearly constant as target visibility falls from all scenes to one in eight. OWLv2 and SAM\,3 retain positive correlations with visibility, but all three detectors lose target-instance localization under heavy occlusion. Grounding DINO reports a hidden target present in about nine of ten scenes by detecting same-category distractors. Consequently, confidence understates the benefit of additional views ($8$ versus $88$ percentage points). In a $40$-scene localization comparison, additional-view selection and fusion increase target localizability from $5\%$ to $100\%$. Experiments cover nine categories, two simulators, and real video. On DAVIS, confidence correlates positively with visible area, yet $99\%$ of occluded frames yield a detection and $94\%$ of those detections miss the tracked target. These results distinguish category presence from target visibility and motivate target-grounded evaluation of active perception.

\end{abstract}

\begin{IEEEkeywords}
active perception, open-vocabulary detection, occlusion, foundation models,
sensing reliability, benchmark
\end{IEEEkeywords}

\section{Introduction}
An open-vocabulary detector maps a text query to regions and confidence scores for the named category. Such scores can also be read as a proxy for whether a particular target is visible. This interpretation assumes that category confidence tracks target visibility. We test that assumption against detector-free geometry, finding the strongest decoupling in Grounding DINO. When we occlude a named target behind a wall of distractors until almost none of it remains visible, the confidence barely moves, stays uncorrelated with the true
visibility measured by a detector-free oracle, and still reports the target present
in most scenes, because a same-category object beside the target keeps the score
high. We establish this decoupling against ground truth and trace its
consequences for the systems that read the signal.

This assumption appears wherever a category score stands in for target visibility:
filtering detections by a confidence threshold, evaluating an open-vocabulary
detector by whether it reports the queried object, grounding a referring expression
to a region, and retrieving a specific instance from a gallery. The assumption is most directly measurable in active perception, where a system reads the confidence to decide whether it still needs to look. A robot told to pick up a named object in clutter
often cannot see it from its default camera, and systems such as
VISO-Grasp~\cite{shi2025viso} and ActiveGrasp~\cite{lei2026activegrasp} move the
camera to reveal the object, deciding when to move and whether a view worked from an
open-vocabulary detector's confidence. We instrument this setting because it supplies
a geometry oracle and a measurable downstream consequence, and isolate the detector signal before it enters a manipulation controller, and motivates testing the same distinction in those other applications. Grounding, retrieval, and complete agentic systems are not directly evaluated here.

The same signal now sits inside agentic sensing systems, where a language-driven
agent decides when to acquire another view. A camera that can be moved, an on-board
detector, and a policy that asks the detector whether the named object is in view
together form a trigger: a high score means act on the current view, a low one means
take a further view. The trigger keeps the number of views small, since every further view means camera motion, added latency, and, on a networked platform, more transmitted frames. In the controlled Grounding DINO experiment, this trigger is inverted. The score is highest in the layouts where a further view is needed most, so
an agent that gates on it acquires views where the target is already visible and reports success from a view without the target.

\begin{figure*}[t]
\centering
\includegraphics[width=0.98\textwidth]{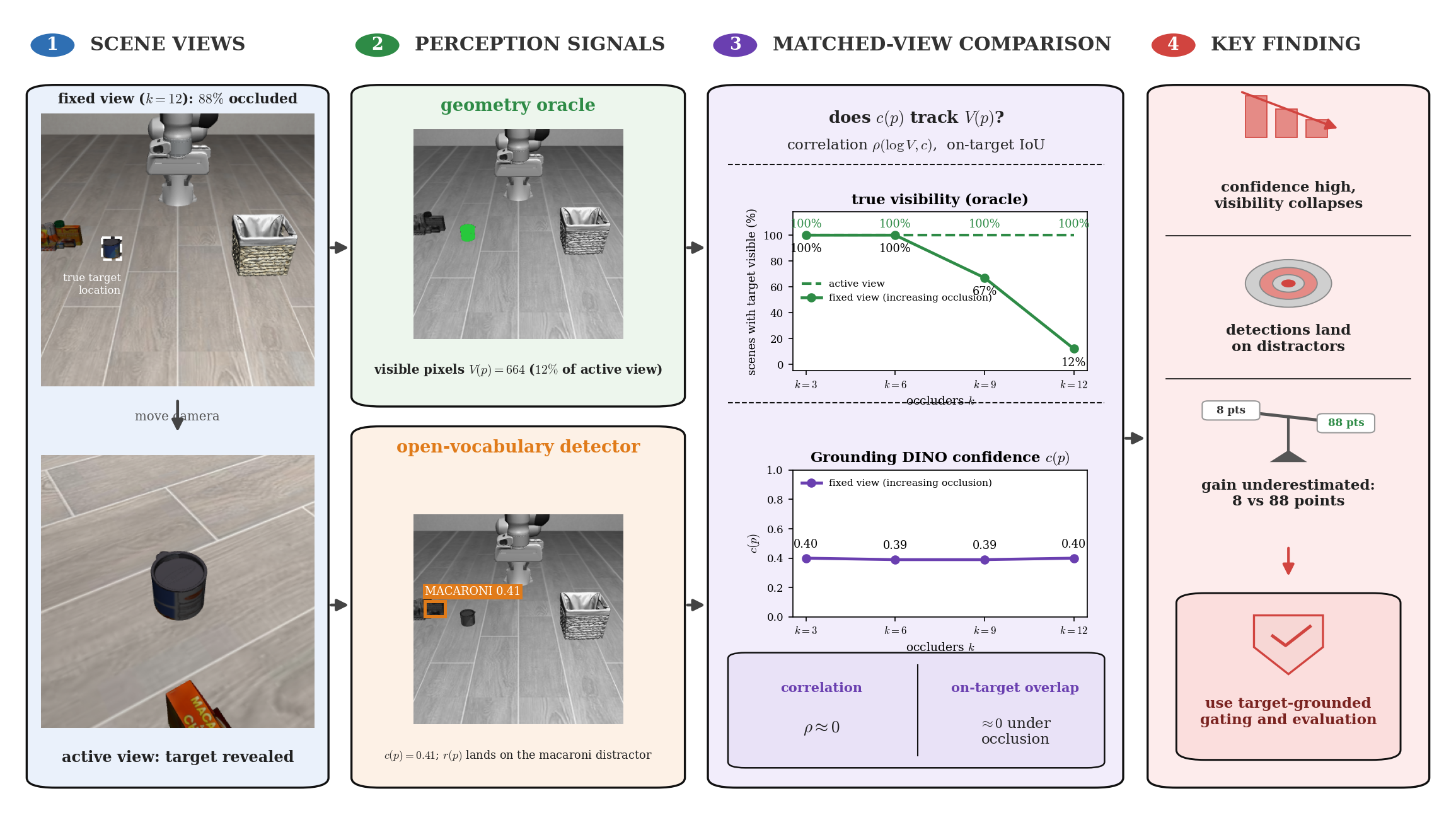}
\caption{Overview of the geometry-oracle evaluation. From each scene (column~1) we
take a fixed default view, where distractors occlude most of the target, and an active
view that reveals it. Two perception signals (column~2) read each view: a
detector-free geometry oracle that counts the target's visible pixels $V(p)$, and an
open-vocabulary detector that returns a confidence $c(p)$ and a region $r(p)$, which
here lands on a distractor. We compare them on matched views (column~3): as
occlusion grows, true visibility collapses ($100\%\!\to\!12\%$) while the detector's
confidence stays flat ($\approx 0.40$). The finding (column~4) is that confidence
signals category presence, not the specific target's visibility, which makes a confidence metric understate the benefit of active perception ($8$ versus $88$ points) and a confidence gate fire when the target is hidden.}
\label{fig:overview}
\end{figure*}

Our study pairs every rendered view with a ground-truth oracle that no detector can
influence, namely the count of target pixels visible in a geometry segmentation of
the scene (Fig.~\ref{fig:overview}). With this oracle we can separate what the object's
geometry actually exposes from what the detector reports. We build scenes in which
distractors form a wall between the camera and the target, so occlusion grows in a
controlled way, and we measure, at the fixed default view and at the best of a set
of candidate views, both the oracle visibility and the detector confidence.

The oracle confirms that occlusion is real and that moving the camera resolves it, while Grounding DINO confidence does not follow. OWLv2 and SAM\,3 retain positive confidence--visibility correlations, but their confident detections also lose overlap with the intended target under heavy occlusion. As the target disappears behind the wall that confidence
barely changes, and the detector reports the target found almost as often as when
the target is fully visible, because it fires on nearby distractors of the same
kind. This decoupling has direct consequences for the two roles the score plays, with instance-localization failures across three detectors of different design.

We make three contributions. First, we introduce a geometry-oracle evaluation, a
detector-free measurement of whether an open-vocabulary detector's confidence tracks
the true visibility of a named object, obtained by comparing it against ground-truth
segmentation on matched views, which any group can run on its own detector. Second,
we use it to document the failure and its consequences: Grounding DINO confidence is uncorrelated with true visibility and reports a hidden target as present in most heavily occluded scenes. That understates the benefit of active perception by about a factor of ten and makes a confidence gate fire precisely when the object is hidden. The failure holds across three detectors under a pixel-level intersection-over-union (IoU) attribution, nine object categories, a
randomized occlusion, a second simulator with a different renderer and object set,
and real video from a standard segmentation benchmark, where the detector reports a
hidden target present in almost every occluded frame and localizes to another
instance. Third, we test additional-view selection and fusion as a localization component. The tested single-view confidence and tracked-location signals are ambiguous when occluders occupy the target's expected image location. Selecting and fusing additional views makes the target localizable in all $40$ evaluated scenes, compared with $5\%$ from a fixed view; this measures the combined acquisition and fusion procedure. A common ring occluder layout, finally, leaves the target visible throughout, so the confidence drops reported on it measure clutter rather than occlusion. We recommend target-grounded evaluation that distinguishes category presence from recovery of the intended instance.

\section{Related Work}
Active perception and next-best-view planning choose informative viewpoints to
improve a downstream task. Classic planners drive exploration from volumetric
information gain over occluded space; the closed-loop planner of
Breyer \emph{et al.}~\cite{breyer2022closed} reconstructs the scene online and decides at each step
whether to grasp or to move for a better view. Recent systems fold foundation
models into this loop. VISO-Grasp~\cite{shi2025viso} combines a large
vision-language model with a geometric next-best-view field and an
open-vocabulary detector for target-driven grasping under severe occlusion, and
reports high success on a real robot. ActiveGrasp~\cite{lei2026activegrasp}
greedily selects views that most reduce the uncertainty of a grasp-distribution
model. These systems establish that active perception improves grasping, and they
judge whether a view found the target through foundation-model detection at that
view. Our study takes that judgment as the object of analysis rather than the
tool, and asks whether the detection signal reflects occlusion at all.

Open-vocabulary detection and segmentation provide the perception signal these
systems rely on. Grounding DINO (GDINO)~\cite{groundingdino} detects objects named by free
text, building on grounded and open-vocabulary detectors such as
GLIP~\cite{glip}, OWL-ViT~\cite{owlvit}, and YOLO-World~\cite{yoloworld}, and
OWLv2~\cite{owlv2} scales this through self-training. The Segment Anything
family~\cite{sam,sam2} provides promptable segmentation, and Segment Anything
Model 3 (SAM\,3)~\cite{sam3} adds a presence score for a short noun phrase. Their
scores are commonly read as a proxy for whether the named object is visible. We
test that reading against a geometry oracle and find that confident matches to
similar distractors dominate once the target is occluded.

Related failures appear outside robotics. Modern neural classifiers and detectors
are miscalibrated, giving confident scores that overstate their
correctness~\cite{guo2017calibration}, and large vision-language models hallucinate
objects, confidently describing things that are absent~\cite{pope}. We document a
spatial, occlusion-specific form of this in manipulation, where the detector stays
confident about a named target precisely because a similar object is visible in its
place.

Robustness to clutter has also been pursued inside the policy rather than the
camera. Concept-gated visual distillation~\cite{cgvd2026} teaches a
vision-language-action model, such as OpenVLA~\cite{openvla}, to attend to
task-relevant objects and ignore distractors. A parallel effort makes the
manipulation policy itself efficient to learn, transferring reinforcement-learning
components from simpler upstream tasks to speed up motion planning on harder
ones~\cite{he2025fewshot,he2026sample}. These lines adapt the policy to its task,
whereas we evaluate the perception signal used to decide when to look and whether
looking worked.

A smaller line of work evaluates foundation models in robotics rather than extending
them. Sui \emph{et al.}~\cite{sui2025grounding} compare integration paradigms and report
that an open-vocabulary detector reaches only $0.3$ to $0.4$ grounding accuracy
under realistic conditions, and ElMallah \emph{et al.}~\cite{elmallah2025score} argue that a single task-success number hides where a policy fails, and instead score per subgoal. Our study shares this stance and narrows it to one signal, the detector
confidence, using a geometry oracle to distinguish visibility sensitivity from target-instance attribution.

\section{Geometry-Oracle Evaluation}
We propose a geometry-oracle evaluation of the perception signal that active-perception
systems use. The idea is to pair every view with a detector-free measurement of
what the scene geometry actually exposes, and to compare it against what the
detector reports on the same view.

\subsection{Problem Formulation}
A scene contains one target object $o^\star$ named by a text query $q$, together
with $k$ distractor objects, where $k$ sets the occlusion severity. A camera at
pose $p$ renders an image $I(p)$. Two functions are defined on this image.

The oracle reads scene geometry. From a segmentation render we obtain the target
mask $M^\star(p)$, whose pixel count $V(p)=|M^\star(p)|$ is the target's true
visibility at pose $p$. No detector influences $V(p)$. We call the target visible
at $p$ when $V(p)\ge v_0$, with $v_0=40$ pixels.

The detector reads appearance. A detector $D$ maps $(I(p),q)$ to a confidence
$c(p)\in[0,1]$ and a predicted region $r(p)$ for its top-scoring detection. We call
the target resolved at $p$ when $c(p)>\tau$, where $\tau$ is a per-detector
threshold.

A strategy selects the pose. The fixed strategy uses the default camera $p_{\text{fix}}$.
The active strategy searches a candidate set $\mathcal{P}$ of viewpoints placed on
a hemisphere over the target and keeps the one that maximizes its own signal, $p_{\text{act}}=\arg\max_{p\in\mathcal{P}} c(p)$ for the detector, or the
analogous choice under $V$ for the oracle. For a signal $s\in\{V,c\}$ we define the
benefit of active perception as the change in the fraction of scenes in which the
target is visible or resolved,
\begin{equation}
B_s \;=\; \Pr\!\big[s\ \text{succeeds at}\ p_{\text{act}}\big]\;-\;\Pr\!\big[s\ \text{succeeds at}\ p_{\text{fix}}\big].
\end{equation}
The question we study is whether the detector confidence $c$ tracks the true
visibility $V$. We measure this with the correlation $\rho(\log V, c)$ across
scenes, and, at the fixed view, with whether a confident detection is placed on the
target, quantified by the overlap $\mathrm{IoU}\!\big(r(p_{\text{fix}}),
M^\star(p_{\text{fix}})\big)>0.5$.

\subsection{Benchmark and Occlusion Control}
We build occluded pick-and-place scenes in the LIBERO simulator. A target object
sits at the center and $k$ distractor food items are added, with
$k\in\{3,6,9,12\}$. Two layouts realize very different occlusions from the same
count. In the frontal layout the $k$ distractors form a wall on the camera side of
the target, so they progressively hide it from the fixed view. In the ring layout
the distractors surround the target symmetrically. We use three target categories,
a soup can, a milk carton, and a tomato-sauce can, each placed at the same position under the same occluder geometry, and draw $N=25$ randomized scenes per target and per $k$.

\subsection{Signals and Attribution}\label{ssec:signals}
For every view we record the oracle visibility $V(p)$ from the segmentation mask,
and the confidence $c(p)$ of each detector, using Grounding DINO at two sizes,
OWLv2, and the SAM\,3 presence score. The active strategy evaluates a set of
$|\mathcal{P}|=24$ candidate views on a hemisphere aimed at the target. To decide
whether a confident detection is actually placed on the target rather than on a
distractor, we compare the predicted region against the ground-truth target mask with a pixel-level overlap: the box against the mask bounding box for the box detectors, and the predicted mask against the target mask for SAM\,3. A detection is on-target when this overlap exceeds one half. We report bootstrap
95\% confidence intervals over scenes with 2000 resamples.

\section{Experiments and Results}
\subsection{Implementation}
We use the LIBERO simulator on robosuite and MuJoCo, rendered at $224\times224$ with
OSMesa. The main experiment uses three target categories (soup, milk, tomato-sauce); the breadth experiment uses nine, with $N=20$. In the frontal layout
the $k$ distractors are placed on the camera side of the target in an arc of radius
$0.035$ to $0.055$\,m with $\pm 6$\,mm jitter; the randomized layout scatters them
through a camera-side cone of half-angle $55^\circ$ and radius $0.03$ to $0.055$\,m;
the ring layout spaces them evenly around the target. The active set $\mathcal{P}$
has $24$ candidate look-at poses ($12$ in the breadth experiment) on a hemisphere over
the target, at azimuths $0^\circ$ to $315^\circ$ in $45^\circ$ steps, elevations
$25^\circ$ and $45^\circ$, and radii $0.35$ and $0.5$\,m.

The detectors are Grounding DINO-tiny and -base
(\texttt{IDEA-Research/grounding-dino-\{tiny,base\}}), OWLv2
(\texttt{google/owlv2-base-patch16-ensemble}), and SAM\,3 (ViT-L), all from public
checkpoints; the confidence $c$ is the maximum detection score, and for SAM\,3 the
presence score. A view is visible when the ground-truth segmentation shows at least
$v_0=40$ target pixels, and resolved when the Grounding DINO score exceeds $\tau=0.35$. For OWLv2 and SAM\,3, whose scores use different scales, we report the correlation with visibility and the mean in place of a thresholded rate, and compare all three at a matched operating point in Section~\ref{ssec:gate}. A detection is
on-target when its overlap with the ground-truth target mask exceeds $0.5$,
box-against-mask-bounding-box for the box detectors and mask-against-mask for SAM\,3.
The confidence gate thresholds $c$; the localization gate uses the distance from the
detection to the target location, taken either from the current-view segmentation
(an upper bound) or from a tracked location from an earlier unoccluded view with
Gaussian tracker noise of standard deviation $\sigma\in\{0,10,20,30\}$ pixels. Gate
quality is the area under the receiver operating characteristic curve for separating
occluded ($V<v_0$) from visible fixed-view scenes, and all confidence intervals are
bootstrap $95\%$ intervals over scenes with $2000$ resamples.

The downstream localization study runs $40$ occluded scenes with a single target
(soup) surrounded by five same-scale distractors that occlude it from the fixed
camera. The localizer back-projects the
SAM\,3 target mask through the metric depth map into the world with the camera
transform, then takes an occlusion-aware estimate: points are restricted to a
table-height band, the densest horizontal cluster is kept, and the height is set to a
nominal grasp height. Move-and-verify fuses the three highest-presence views; the
confidence gate uses the fixed view only. A location counts as on-target within a radius of the ground-truth object position, which enters only as a score and never drives the estimate.

For the second simulator we use ManiSkill\,3~\cite{maniskill3} on SAPIEN with a
Vulkan renderer, the PickClutterYCB task at $256\times256$, over $300$ random
layouts; the target is the task's target object, prompted by its Yale-CMU-Berkeley (YCB) category name,
with its mask read from the per-actor segmentation. The real-image experiment uses the Densely Annotated Video Segmentation benchmark DAVIS-2017~\cite{davis2017}, tracking the primary annotated object as the
visibility oracle and prompting with the object category, over forty clips and
$2898$ frames; a frame counts as occluded when the object's visible area falls below
$30\%$ of its per-clip median.

\subsection{The Decoupling of Confidence from True Visibility}
\begin{figure}[t]
\centering
\includegraphics[width=0.95\columnwidth]{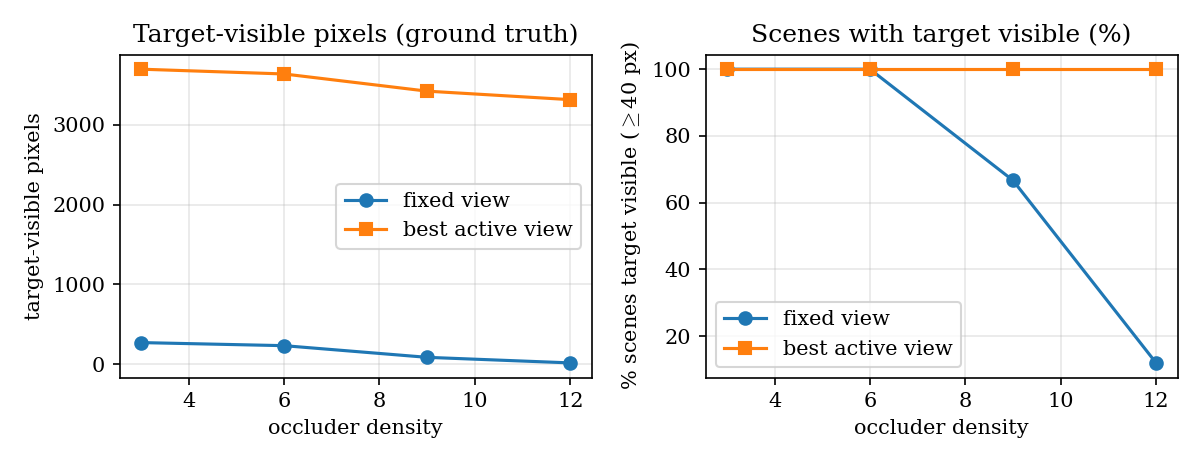}
\caption{Detector-free view of the same scenes. Left, the number of target pixels
visible at the fixed and best active view as occlusion grows; the fixed view loses
the target while the active view keeps it. Right, the fraction of scenes in which
the target is truly visible; it collapses for the fixed view and stays complete for
the active view. Grounding DINO confidence in Table~\ref{tab:decouple} does not
follow the falling fixed-view curve.}
\label{fig:gtvis}
\end{figure}

The oracle confirms that the frontal layout occludes the target and that moving the
camera recovers it (Table~\ref{tab:oracle}, Fig.~\ref{fig:gtvis}). At the fixed
view the fraction of scenes in which the target is actually visible falls from
every scene at $k=3$ and $k=6$ to two in three at $k=9$ and about one in eight at
$k=12$. The best of the $24$ candidate views keeps the target visible in every scene. The oracle benefit of active perception therefore reaches $88$ points under heavy occlusion, so moving the camera recovers the target, and the gain from moving grows with occlusion.

\begin{table}[t]
\centering
\footnotesize
\setlength{\tabcolsep}{3pt}
\caption{Visible-target frequency at the fixed view and the best oracle-selected view. Benefit is the difference in percentage points; brackets give scene-bootstrap $95\%$ confidence intervals ($2000$ resamples).}
\label{tab:oracle}
\begin{tabular}{lcccc}
\toprule
occluders $k$ & $3$ & $6$ & $9$ & $12$ \\
\midrule
visible, fixed view  & $100\%$ & $100\%$ & $67\%$ & $12\%$ \\
visible, active view & $100\%$ & $100\%$ & $100\%$ & $100\%$ \\
benefit (points)     & $0$ & $0$ & $33\,[23,44]$ & $88\,[81,95]$ \\
\bottomrule
\end{tabular}
\end{table}

Grounding DINO's confidence barely responds to this occlusion (Table~\ref{tab:decouple}). As true
visibility falls from every scene to one in eight, the mean confidence of Grounding
DINO stays near $0.40$ and its resolved rate stays above $90\%$. Grounding DINO-tiny behaves the same way at the fixed view, with a mean confidence of $0.48$ at $k=3$ against $0.47$ at $k=12$ and a resolved rate that never falls below $89\%$. Per target, the correlation between true visibility and
confidence is close to zero, between $-0.22$ and $+0.21$. The detector reports the target present almost independently of whether the target is visible.

Region attribution identifies the source of these detections. At the highest density, the target is below $v_0$ in $88\%$ of scenes, and Grounding DINO still exceeds its confidence threshold in $91\%$ of that subset. The fraction of confident detections overlapping the intended target falls from $29\%$ with the target fully visible to $0\%$ under heavy occlusion. The remaining confident boxes lie on distractors of the same category.

\begin{table}[t]
\centering
\footnotesize
\setlength{\tabcolsep}{3pt}
\caption{Fixed-view visibility and Grounding DINO-base scores. Change is from $k=3$ to $k=12$, in percentage points for frequencies and score units for confidence. Values are rounded independently.}
\label{tab:decouple}
\begin{tabular}{lccccc}
\toprule
occluders $k$ & $3$ & $6$ & $9$ & $12$ & change $3{\to}12$ \\
\midrule
visible (oracle) & $100\%$ & $100\%$ & $67\%$ & $12\%$ & $-88$ pts \\
confidence       & $0.40$ & $0.39$ & $0.39$ & $0.40$ & $-0.01$ \\
resolved rate    & $95\%$ & $97\%$ & $95\%$ & $92\%$ & $-3$ pts \\
\bottomrule
\end{tabular}
\end{table}

The first consequence is a mismeasurement. Because fixed-view confidence is already
high through false positives, the benefit of active perception measured through
confidence is small and flat, from $3$ to $8$ points for Grounding DINO-base across
$k$, reading $8\,[3,15]$ at the highest density against the oracle's $88\,[81,95]$. A confidence-based evaluation therefore
understates the benefit of moving the camera by about a factor of ten, and would
conclude that active perception barely helps in exactly the regime where it helps
most.

The second consequence is a broken gate. A system that moves the camera only when
the detector is unsure will rarely move, since the detector exceeds its threshold in $92\%$
of scenes at the highest occluder density. The gate reports the
target visible and stays put while the target is occluded, which is the worst
failure for a trigger meant to decide when to look.

\subsection{The Limits of Single-View Gating Signals}\label{ssec:gate}
We next locate the boundary of what one view can support. We treat the fixed-view decision as a binary problem, occluded ($V<v_0$)
against visible, and score signals by the area under the receiver operating
characteristic curve, where $0.5$ is chance (Table~\ref{tab:gate}). Confidence sits
at chance for Grounding DINO, $0.50\,[0.42,0.57]$. OWLv2 separates the two classes better, $0.76\,[0.71,0.81]$, because its score does fall under occlusion, but only into a range where a threshold is fragile, $0.12$ unoccluded against $0.04$ under heavy occlusion; SAM\,3 presence stays at chance. A
localization signal, whether the detection overlaps the target, does separate the
two well, up to $0.86$ for OWLv2, but only when it is computed against the target's
segmentation in the current view. That segmentation is exactly what occlusion
removes, so this number is an upper bound a system cannot compute before it moves, not a usable gate. When we replace it with a realizable signal, the distance from
the detection to the target's location tracked from an earlier unoccluded view, the
gate collapses. Grounding DINO stays at chance ($0.50$), and OWLv2 reaches only
$0.72$ with perfect memory of the location and falls to chance once that location carries a modest error, $0.53$ at ten pixels, $0.53$ at twenty, and $0.45$ at thirty. The reason is geometric: the occluding
distractors sit directly in front of the target, so a detection on an occluder is
close to where the target is, and a location check cannot tell the two apart. These tested single-view signals remain ambiguous when an occluder occupies the target's expected image location. The result supports acquiring another view in this layout; it does not establish an impossibility result for every single-view estimator.

Because the three scores live on different scales, we also compare them at a matched
operating point: each threshold is the fifth percentile of that detector's scores on
the scenes where the oracle says the target is visible, so all three keep the same
share of genuinely visible views (Table~\ref{tab:matched}). There Grounding DINO still
reports the target resolved in $85\%$ of the most heavily occluded scenes and OWLv2 in
$95\%$, so the detector whose confidence carried the most information about visibility
still produces a high false-positive rate at this high-recall operating point. SAM\,3 has $221$ of $300$ presence scores at a shared floor of $0.05$; the threshold shown retains both visible and occluded cases. Thus, a positive association with visibility does not by itself guarantee target-grounded decisions at a selected threshold.

\begin{table}[t]
\centering
\footnotesize
\setlength{\tabcolsep}{3pt}
\caption{Single-view gate AUROC for occluded versus visible scenes. Current-view target segmentation supplies an oracle upper bound; tracked location uses an earlier unoccluded view. SAM\,3 presence and mask-overlap AUROC are $0.46$ and $0.57$, respectively.}
\label{tab:gate}
\begin{tabular}{lcc}
\toprule
gate signal & GDINO & OWLv2 \\
\midrule
confidence                          & $0.50$ & $0.76$ \\
localization, current-view segmentation (upper bound) & $0.68$ & $0.86$ \\
localization, tracked location (realizable)           & $0.50$ & $0.72$ \\
\bottomrule
\end{tabular}
\end{table}

\begin{table}[t]
\centering
\footnotesize
\setlength{\tabcolsep}{4pt}
\caption{Matched high-recall operating point. Thresholds are the fifth percentile of scores on oracle-visible scenes. Tied SAM\,3 scores at $0.05$ retain all scenes. Resolved rates count category detections, independently of target overlap.}
\label{tab:matched}
\begin{tabular}{lcccc}
\toprule
detector & threshold & visible kept & resolved $k{=}3$ & resolved $k{=}12$ \\
\midrule
GDINO-base & $0.354$ & $95\%$ & $92\%$ & $85\%$ \\
OWLv2      & $0.003$ & $95\%$ & $100\%$ & $95\%$ \\
SAM\,3     & $0.050$ & $100\%$ & $100\%$ & $100\%$ \\
\bottomrule
\end{tabular}
\end{table}

The geometry and detector measurements answer complementary questions. Visible-target frequency asks whether sufficient target pixels are exposed, while region attribution asks whether the detector actually selects those pixels. A category can be detected correctly somewhere in the image even when the intended target is not available for manipulation. Separating these measurements prevents a high-scoring distractor from being counted as recovery of the target. It also allows view acquisition to be judged independently of the detector used to report success.

This separation is particularly relevant when choosing an operating threshold. Lowering the threshold may increase recall of category matches, but cannot by itself establish which physical instance generated a detection. Conversely, a detector whose confidence falls with occlusion can still produce regions on another instance. We therefore retain both confidence--visibility association and target-overlap measurements in the cross-detector comparison below. The two characterize different aspects of sensing reliability and need not improve together.

\subsection{Generality across Detectors, Objects, Simulators, and Real Video}
The effect appears in detectors of unrelated design. We repeat the fixed-view
measurement with OWLv2, a detector trained with image-text contrastive learning, and
SAM\,3, a promptable segmentation model, and we score attribution with the pixel-level overlap of Section~\ref{ssec:signals}. For all three detectors the fraction of confident detections that land on the target falls to near zero under heavy occlusion (Table~\ref{tab:generality}), so none of them localizes the occluded target even when it fires.
Grounding DINO is the extreme case, since its confidence is flat and uncorrelated with visibility and it stays confident while wrong. Its top-scoring box overlaps the target's box by $0.32$ on average even with the target fully visible, against $0.55$ for OWLv2, because that box tends to enclose the whole cluster rather than the named can. OWLv2 and SAM\,3 partly track
visibility, with correlations of $0.45\,[0.38,0.51]$ and $0.25\,[0.17,0.32]$, compared with $0.00\,[-0.12,0.14]$ for Grounding DINO. Their scores decrease as occlusion increases. This correlation is useful information, but the matched-threshold results in Table~\ref{tab:matched} show that high recall can still coincide with frequent detections of hidden targets. These independently drawn layouts reproduce Grounding DINO confidence to within $0.004$ of Table~\ref{tab:decouple}.

\begin{table}[t]
\centering
\scriptsize
\setlength{\tabcolsep}{2pt}
\caption{Fixed-view comparison across detectors. Correlation, mean confidence at each density, and on-target detections (overlap $>0.5$) distinguish visibility sensitivity from instance localization. Grounding DINO is nearly uncorrelated with visibility; OWLv2 and SAM\,3 retain positive correlations.}
\label{tab:generality}
\begin{tabular}{lcccccc}
\toprule
detector & $\rho(c,V)$ & $c_{k=3}$ & $c_{k=6}$ & $c_{k=9}$ & $c_{k=12}$ & on-target $k{=}3{\to}12$ \\
\midrule
GDINO-base & $+0.00$ & $0.40$ & $0.39$ & $0.39$ & $0.39$ & $29\% {\to} 0\%$ \\
OWLv2      & $+0.45$ & $0.12$ & $0.10$ & $0.10$ & $0.04$ & $100\% {\to} 0\%$ \\
SAM\,3     & $+0.25$ & $0.25$ & $0.11$ & $0.20$ & $0.05$ & $100\% {\to} 0\%$ \\
\bottomrule
\end{tabular}
\end{table}

The effect extends beyond three objects and beyond a placed wall. We repeat
the measurement on nine target categories with a randomized frontal occlusion, in which
the distractors are scattered through the camera-side cone in front of the target
rather than placed in an arc (Table~\ref{tab:breadth}). The oracle benefit of active
perception again grows with density, to $+84$ points, so the occlusion is real and
moving resolves it. Grounding DINO again does not follow the oracle: as true visibility falls from $52\%$ to $16\%$ of scenes, its confidence edges up from $0.48$ to $0.51$ and its resolved rate reaches $100\%$. Within each of the nine categories
the correlation between confidence and true visibility stays near zero, between
$-0.27$ and $+0.16$, and pooling the categories after removing each category's mean leaves
$-0.09\,[-0.17,0.00]$ over $720$ scenes, so the decoupling is uniform across
categories rather than carried by any one of them. The per-object oracle benefit of
moving ranges from $+12$ to $+100$ points.

\begin{table}[t]
\centering
\footnotesize
\setlength{\tabcolsep}{3pt}
\caption{Nine categories under randomized frontal occlusion ($720$ scenes). Oracle benefits are in percentage points, with $95\%$ intervals $[41,56]$, $[62,76]$, $[76,87]$, and $[78,89]$ across the four densities.}
\label{tab:breadth}
\begin{tabular}{lcccc}
\toprule
occluders $k$ & $3$ & $6$ & $9$ & $12$ \\
\midrule
visible, fixed (oracle)  & $52\%$ & $31\%$ & $19\%$ & $16\%$ \\
visible, active (oracle) & $100\%$ & $100\%$ & $100\%$ & $100\%$ \\
oracle benefit (points)  & $+48$ & $+69$ & $+81$ & $+84$ \\
confidence               & $0.48$ & $0.50$ & $0.51$ & $0.51$ \\
resolved rate            & $94\%$ & $97\%$ & $99\%$ & $100\%$ \\
\bottomrule
\end{tabular}
\end{table}

Table~\ref{tab:perobject} reads the same measurement one category at a time. Every category carries the effect on its own, and the level of the score follows what the object looks like rather than how much of it the camera can see: cream cheese, visible from the fixed camera in none of its scenes, scores $0.60$, while ketchup, visible in $88\%$ of its scenes, scores $0.39$. Moving the camera raises the fraction of scenes with the target visible by $12$ to $100$ points, and the score barely registers the difference.

\begin{table}[t]
\centering
\scriptsize
\setlength{\tabcolsep}{4pt}
\caption{Per-category results pooled over four densities ($80$ scenes per category). Visibility and benefit are frequencies and percentage-point differences, respectively. Confidence remains weakly associated with target visibility within each category.}
\label{tab:perobject}
\begin{tabular}{lccccc}
\toprule
target & $\rho(c,V)$ & visible, fixed & benefit & $c_{k=3}$ & $c_{k=12}$ \\
\midrule
alphabet soup     & $-0.18$ & $12\%$ & $+88$  & $0.44$ & $0.45$ \\
bbq sauce         & $-0.27$ & $21\%$ & $+79$  & $0.45$ & $0.53$ \\
butter            & $+0.16$ & $0\%$  & $+100$ & $0.53$ & $0.55$ \\
chocolate pudding & $-0.09$ & $1\%$  & $+99$  & $0.57$ & $0.62$ \\
cream cheese      & $-0.01$ & $0\%$  & $+100$ & $0.60$ & $0.61$ \\
ketchup           & $+0.12$ & $88\%$ & $+12$  & $0.39$ & $0.38$ \\
milk              & $-0.22$ & $72\%$ & $+28$  & $0.39$ & $0.40$ \\
orange juice      & $-0.10$ & $50\%$ & $+50$  & $0.50$ & $0.54$ \\
tomato sauce      & $+0.00$ & $19\%$ & $+81$  & $0.46$ & $0.50$ \\
\bottomrule
\end{tabular}
\end{table}

The effect also crosses simulators, renderers, and object libraries. We replicate
the measurement in ManiSkill~\cite{maniskill3}, which runs on the
SAPIEN engine~\cite{sapien} with a Vulkan renderer rather than MuJoCo with OSMesa,
and draws on YCB household objects rather than food cans. We use the PickClutterYCB
task, in which several YCB objects sit in natural clutter and occlude one another,
so the occlusion comes from the scene rather than from a wall we place. Over $300$
random layouts the finding holds. Grounding DINO's confidence is uncorrelated with
the target's ground-truth visibility ($\rho=0.08\,[-0.05,0.20]$), and its mean barely moves between
the least and the most visible quartile of layouts, $0.44$ against $0.47$. What moves
is where the box goes: on the least visible quartile it reports the target present in $75\%$ of cases with the box on the target in $3\%$, against $36\%$ on the most visible quartile, so correct localization falls by an order of magnitude while the score holds.

To see whether the effect survives outside simulation we run the same measurement on real
video from DAVIS-2017~\cite{davis2017}, using each clip's ground-truth mask as the
visibility oracle and the object category as the prompt, over $2898$ frames of
forty clips. On real footage the identity failure is at its most pronounced. Confidence keeps a positive correlation with true visibility, $\rho=0.29$ pooled, and its mean falls from $0.73$ when the object is fully visible to $0.53$ once the object shrinks below a third of its typical size. Yet on the $138$ occluded frames it reports the object present in $99\%$ of cases ($[98,100]$ with the clip as the resampling unit), and in $94\%$ of those its box lies off the target. Splitting the clips isolates the mechanism. In the three clips that hold another instance of the queried category the box never lands on the tracked object, across all $39$ of their occluded frames, while in the six single-object clips it lands on it in $8\%$ of $99$ frames. A score that moves with visibility is still not a report about the tracked instance: it says the category is present somewhere, and its box goes to another instance of that category once the target is hidden. DAVIS varies viewpoint, apparent scale, background, and appearance as well as occlusion, whereas the controlled simulation fixes the scene construction. Those additional variations offer a plausible explanation for the positive pooled confidence--visibility correlation. The real-video result therefore supports the distinction between category detection and tracked-instance localization, while the controlled experiment isolates the response to added occluders.

\subsection{Layout Geometry and Threshold Controls}
The symmetric ring layout, common in cluttered manipulation benchmarks, misleads in a different way. With the same $k$ distractors placed in a ring rather than a frontal
wall, the oracle shows the target stays visible from the fixed camera at every density, near $209$ pixels, in every scene up to $k=9$ and in all but one at $k=12$, because the ring never blocks the line of sight. Yet the detector's resolved rate there falls from every scene to about $40\%$ as the ring grows. That drop measures clutter, the
detector losing confidence among many similar objects, not occlusion. Put together
with the frontal result, confidence stays high when the target is truly occluded and
drops when it is merely surrounded, so it tracks neither. A benchmark that labels a
ring of distractors as occlusion and scores it with detector confidence measures two
different things at once.

\begin{figure}[t]
\centering
\includegraphics[width=0.95\columnwidth]{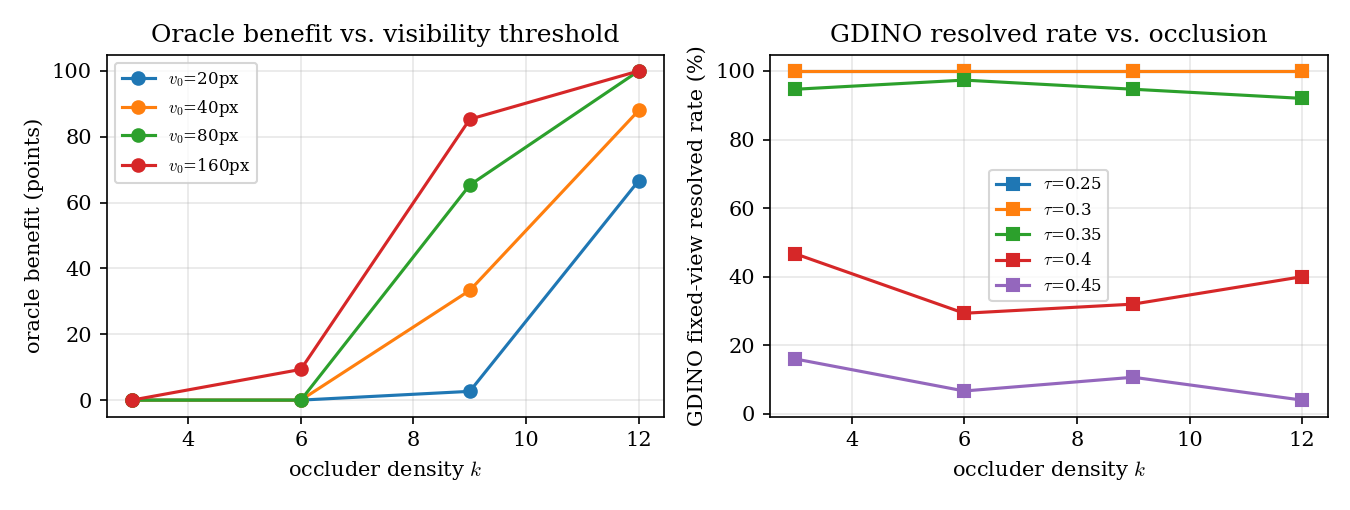}
\caption{Robustness to thresholds. Left, the oracle benefit of active perception
across occluder density, for a range of visibility thresholds $v_0$; the shape is
preserved. Right, Grounding DINO's fixed-view resolved rate across occluder density,
for a range of confidence thresholds $\tau$. Over the range that keeps the detector
usable on unoccluded scenes, $\tau\le0.35$, the rate stays flat and never follows the
collapse in true visibility; the two higher thresholds resolve only $47\%$ and $16\%$
of unoccluded scenes to begin with.}
\label{fig:robust}
\end{figure}

The two headline effects do not depend on the particular thresholds we chose
(Fig.~\ref{fig:robust}). The oracle benefit of active perception keeps the same
shape across occluder density for visibility thresholds from $20$ to $160$ pixels,
staying near zero under light clutter and large under heavy occlusion. In parallel,
Grounding DINO's fixed-view resolved rate stays flat across occlusion at every
threshold that keeps the detector usable on unoccluded scenes, $\tau$ from $0.25$ to
$0.35$, where it resolves $95\%$ of them or more. Across the operating range a system
would use, the decoupling of confidence from visibility holds.

\subsection{The Downstream Consequence for Target Localization}
We next compare the world position supplied to a planner in $40$ occluded scenes. The confidence gate uses the fixed high-confidence view (Grounding DINO $0.44$). Move-and-verify searches candidate views and fuses the three with highest target-presence scores. Both use the SAM\,3 mask-and-depth localizer described above, with ground truth used to score positions. The comparison measures the combined effect of additional-view selection and fusion on localization.

The gap is at the level of whether the target can be localized at all
(Table~\ref{tab:closedloop}). From the fixed high-confidence view the target is
localizable in $5\%\,[0,12]$ of scenes, because it is occluded; moving first makes it
localizable in every scene, $100\%\,[100,100]$, and brings the fused position within $10$\,cm of the true
object in $58\%$ of them against $2\%$ from the fixed view. The confidence gives no
warning of the difference. It reads $0.44$ at the fixed view where the target cannot be localized and rises to $0.62$ after moving, so a planner gating on the score would commit to a view from which the target cannot be found while the score reads high.

The residual error has a geometric origin, which a round-trip check isolates. Projecting the ground-truth center to a pixel and back-projecting its depth returns the position exactly in the
horizontal image axis and errs vertically, because an occluder covers the target's center pixel, so the depth read there is the occluder's, while the mask centroid is pulled off by the target's partial visibility. The occlusion and the same-category confusion that decouple the confidence therefore also govern what a mask-plus-depth estimate can recover from a view without the target. This is why a high fixed-view score is not a cue to act: it carries no information about whether the position a planner receives came from a view that contained the target at all.

\begin{table}[t]
\centering
\footnotesize
\setlength{\tabcolsep}{3pt}
\caption{Target localization in $40$ occluded scenes. The confidence gate uses one fixed view; move-and-verify selects and fuses additional views using the same mask-and-depth localizer. Bold marks the better value. A median localization error is unavailable for the confidence gate.}
\label{tab:closedloop}
\begin{tabular}{lcc}
\toprule
 & Conf. gate (no move) & Move-and-verify \\
\midrule
GDINO confidence & $0.44$ & $0.62$ \\
Target localizable & $5\%$ & $\mathbf{100\%}$ \\
Loc. within $8$\,cm & $0\%$ & $\mathbf{22\%}$ \\
Loc. within $10$\,cm & $2\%$ & $\mathbf{58\%}$ \\
Median loc. error (m) & n/a & $0.098$ \\
\bottomrule
\end{tabular}
\end{table}

\section{Discussion}
Category detection and target-instance localization require different evidence. A high category score can remain correct about the scene while its region misses the object specified by the task. Grounding DINO exhibits the strongest confidence--visibility decoupling in the controlled scenes. OWLv2 and SAM\,3 retain sensitivity to visibility, but their regions also lose the hidden target. This distinction connects detector calibration~\cite{guo2017calibration} and object hallucination~\cite{pope} to the spatial attribution needed for manipulation.

The practical recommendation is to evaluate view acquisition against target-grounded visibility, region overlap, or world-position error. Such signals measure whether the intended instance was recovered. The geometry oracle supplies this reference in simulation; annotated masks provide a corresponding reference in video. A ring of distractors should be evaluated separately from a frontal occluder because the former can lower confidence while leaving the target visible.

The localization comparison is a tested component of a remedy, rather than a complete active-manipulation controller. It combines candidate-view selection, segmentation, metric depth, and multi-view fusion. An online realization incurs image acquisition, detector inference, camera travel and settling, and depth-fusion costs for each acquired view. Exhaustive evaluation requires up to $24$ candidate views in the main benchmark ($12$ in the breadth study), compared with one fixed view. Fusing three selected views adds reconstruction and registration work. Wall-clock inference time, camera-motion latency, energy use, and end-to-end robot success are not measured here. These costs must be evaluated jointly with localization accuracy when replacing an exhaustive scan with a practical view policy.

The evidence spans controlled same-category interference, randomized layouts, natural YCB clutter, and real annotated video. It most directly characterizes category-to-instance confusion under occlusion. More varied object geometry, different-category occluders, dynamic scenes, and repeated physical-robot trials would establish the boundary in broader manipulation settings. DAVIS visible area also changes with apparent scale and viewpoint, so it is a proxy for occlusion rather than a geometric fraction of the whole object. The direct downstream evidence concerns localization; extensions to language grounding, retrieval, or complete agentic systems remain hypotheses for task-specific evaluation.

\section{Conclusion}
Grounding DINO confidence can remain high as a named target disappears behind clutter. OWLv2 and SAM\,3 respond more to visibility, yet all three can detect a category without localizing the intended instance. In the controlled benchmark, confidence measures an $8$-point active-view benefit where the geometry oracle measures $88$ points. Additional-view selection and fusion improve target localizability from $5\%$ to $100\%$ in the $40$-scene comparison. The ring control shows that confidence can also fall while the target remains visible, separating clutter from occlusion. These findings support evaluation with target visibility and region attribution, alongside acquisition and computation costs when additional views are used.

\balance
\bibliographystyle{IEEEtran}
\bibliography{refs}

\end{document}